\pdfoutput=1

\documentclass[11pt]{article}

\usepackage[final]{EMNLP2022}

\usepackage{times}
\usepackage{latexsym}

\usepackage[T1]{fontenc}

\usepackage[utf8]{inputenc}

\usepackage{microtype}

\usepackage{inconsolata}

\usepackage{beamerarticle}
\usepackage{latexsym}
\usepackage{microtype}
\usepackage{booktabs}
\usepackage{url}
\usepackage{eurosym}
\usepackage{todonotes}
\usepackage{multirow}
\usepackage{subcaption}
\usepackage{amssymb}
\usepackage{amsmath}
\usepackage{mathtools}
\usepackage{array}
\usepackage{pgfplotstable}
\usepackage{pgfplots}
\usepackage{soul}
\usepackage{xcolor}
\usepackage{txfonts}
\usepackage{pgf}
\usepackage{tikz}
\usepackage{adjustbox}
\usepackage{appendix}
\usepackage{enumitem}
\usepackage{longtable}
\usepackage{braket}
\usepackage{makecell}
\definecolor{lightsalmon}{HTML}{FAE1DD}
\definecolor{darksalmon}{HTML}{fec5bb}
\definecolor{lightbeige}{HTML}{FFE6CC}
\definecolor{lightgreen}{HTML}{d8e2dc}
\definecolor{darkorange}{rgb}{0.99,0.67,0.3}
\definecolor{lightorange}{rgb}{0.99,0.89,0.79}
\definecolor{myteal}{rgb}{0,0.5,0.5}
\definecolor{lightteal}{rgb}{0.6,0.77,0.74}
\definecolor{lightmagenta}{rgb}{0.84,0,0.45}
\newcommand{\hllightsalmon}[1]{{\sethlcolor{lightsalmon}\hl{#1}}}

\newcommand{\hllightgreen}[1]{{\sethlcolor{lightgreen}\hl{#1}}}
\newcommand{\hllightbeige}[1]{{\sethlcolor{lightbeige}\hl{#1}}}
\newcommand{\hldarksalmon}[1]{{\sethlcolor{darksalmon}\hl{#1}}}

\newcolumntype{P}[1]{>{\centering\arraybackslash}p{#1}}
\newcolumntype{M}[1]{>{\centering\arraybackslash}m{#1}}
\newcolumntype{L}[1]{>{\PreserveBackslash\raggedright}p{#1}}

\usetikzlibrary{decorations.pathmorphing,patterns}
\usepgfplotslibrary{groupplots,units}
\usetikzlibrary{backgrounds,positioning,fit,shapes}
\pgfplotsset{compat=newest}

\newenvironment{myfont}{\fontfamily{pcr}\selectfont}{\par}

\usepackage{pifont}
\newcommand{\cmark}{\ding{51}}
\newcommand{\xmark}{\ding{55}}

\newif\ifcomments
\commentstrue 
\ifcomments
    \providecommand{\sameer}[1]{{\protect\color{magenta}{[sameer: #1]}}}
    \providecommand{\matt}[1]{{\protect\color{teal}{[Matt: #1]}}}
    \providecommand{\dheeru}[1]{{\protect\color{olive}{[Dheeru: #1]}}}
\else
    \providecommand{\matt}[1]{}
    \providecommand{\sameer}[1]{}
    \providecommand{\dheeru}[1]{}
\fi

\title{Successive Prompting for Decomposing Complex Questions}

\def\authspace{\hspace{5mm}}
\author{
Dheeru Dua\textsuperscript{$\clubsuit$}
\authspace{} 
Shivanshu Gupta\textsuperscript{$\clubsuit$}
\authspace{} 
\textbf{Sameer Singh}\textsuperscript{$\clubsuit, \spadesuit$}
\authspace{} 
\textbf{Matt Gardner}\textsuperscript{$\heartsuit$} 
\\
  \textsuperscript{$\clubsuit$}University of California, Irvine, USA 
  \authspace{} 
  \textsuperscript{$\spadesuit$}Allen Institute for Artificial Intelligence \\
  \authspace{} 
  \textsuperscript{$\heartsuit$}Microsoft Semantic Machines \\
  \texttt{\makecell{\{ddua,shivag5,sameer\}@uci.edu, mattgardner@microsoft.com\\}}}

\begin{document}
\maketitle
\begin{abstract}

Answering complex questions that require making latent decisions is a challenging task, especially when limited supervision is available.  
Recent works leverage the capabilities of large language models (LMs) to perform complex question answering in a few-shot setting by demonstrating how to output intermediate rationalizations while solving the complex question in a single pass.  We introduce ``Successive Prompting'', where we iteratively break down a complex task into a simple task, solve it, and then repeat the process until we get the final solution. Successive prompting decouples the supervision for decomposing complex questions from the supervision for answering simple questions, allowing us to (1) have multiple opportunities to query in-context examples at each reasoning step (2) learn question decomposition separately from question answering, including using synthetic data, and (3) use bespoke (fine-tuned) components for reasoning steps where a large LM does not perform well. 
The intermediate supervision is typically manually written, which can be expensive to collect. We introduce a way to generate a synthetic dataset which can be used to bootstrap a model's ability to decompose and answer intermediate questions. Our best model (with successive prompting) achieves an improvement of $\sim$5\% absolute F1 on a few-shot version of the DROP dataset when compared with a state-of-the-art model with the same supervision.

\end{abstract}

\section{Introduction}

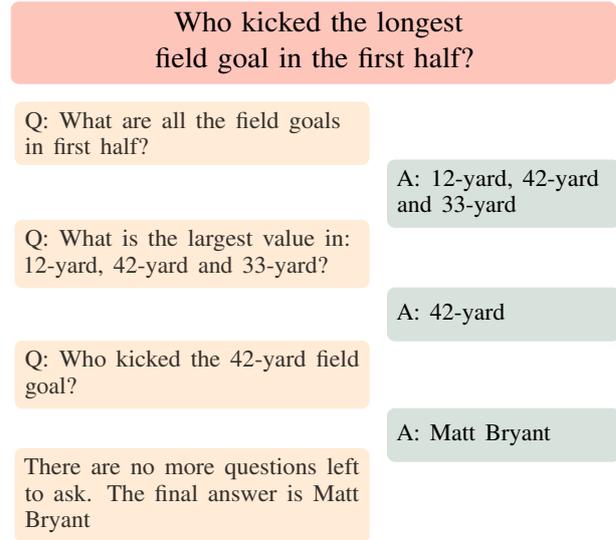
\begin{figure}
  \begin{subfigure}[b]{\columnwidth}
    \begin{tikzpicture}
     \tikzstyle{ellipsis}=[circle,fill=none,text=black,minimum size=0.1cm]
     \tikzstyle{header}=[rectangle,fill=none,text=black,minimum size=0.1cm,text width=2cm]
      \tikzstyle{complex}=[rectangle,draw,darksalmon,fill=darksalmon,align=center,text=black,text width=7.7cm,minimum size=1cm, rounded corners=0.1cm, opacity=1]
      \tikzstyle{simple1}=[rectangle,draw,lightbeige,fill=lightbeige,opacity=0.8,text=black,text width=4.4cm,minimum size=0.5cm, rounded corners=0.1cm]
      \tikzstyle{simple2}=[rectangle,draw,lightbeige,fill=lightbeige,opacity=0.8,text=black,text width=4.4cm,minimum size=0.5cm, rounded corners=0.1cm]
      \tikzstyle{simple3}=[rectangle,draw,lightbeige,fill=lightbeige,opacity=0.8,text=black,text width=4.4cm,minimum size=0.5cm, rounded corners=0.1cm]
       \tikzstyle{answer}=[rectangle,draw,lightgreen,fill=lightgreen,opacity=1,text=black,text width=2.8cm,minimum size=0.7cm, rounded corners=0.1cm]
      
      \node[ellipsis]           (C)                                 {};
      \node[complex]           (CQ) [below of=C,yshift=4.5cm]       {\normalsize{ Who kicked the longest field goal in the first half?}\par};
      \node[simple1]           (S0) [below of=CQ,yshift=-0.2cm,xshift=-1.6cm]       {\footnotesize{Q: What are all the field goals in first half?}\par};
      \node[answer]           (A0) [below of=S0,xshift=4.1cm,yshift=0.2cm]       {\footnotesize{A: 12-yard, 42-yard and 33-yard}\par};
      \node[simple2]           (S1) [below of=S0,yshift=-0.6cm]       { \footnotesize{Q: What is the largest value in: 12-yard, 42-yard and 33-yard?}\par};
       \node[answer]           (A1) [below of=A0,yshift=-0.6cm]       {\footnotesize{A: 42-yard}\par};
        \node[simple2]           (S2) [below of=S1,yshift=-0.6cm]       { \footnotesize{Q: Who kicked the 42-yard field goal?}\par};
       \node[answer]           (A2) [below of=A1,yshift=-0.6cm]       {\footnotesize{A: Matt Bryant}\par};
        \node[simple3]           (S3) [below of=S2,yshift=-0.6cm]       { \footnotesize{There are no more questions left to ask. The final answer is Matt Bryant}\par};
    \end{tikzpicture}
  \end{subfigure}
  \caption{Example decomposition used by Successive Prompting's \hllightbeige{question decomposition} and \hllightgreen{question answering} stage on a DROP example.  The model iterates between predicting a simple question to ask and answering the simple question.}
 \label{fig:sim_q}
\end{figure}

Compositional reading comprehension datasets like HotpotQA~\cite{yang2018hotpotqa} and DROP~\cite{dua2019drop} have inspired a range of model architectures that learn to answer complex questions with weak supervision from the final answer. One recent direction is to leverage large language models (LMs) to solve compositional tasks with very few examples by generating latent reasoning steps before answering the question~\cite{wei2022chain,nye2021show,karpas2022mrkl}. 
Given a complex question, this approach first finds nearest-neighbor training examples from a dataset of (question, reasoning, answer) triples and then concatenates them to create an input for the LM. A large LM is then prompted with this input to generate the intermediate reasoning steps needed, while answering the complex question in a single pass.

While promising, this approach discards many of the benefits of prior approaches to this task~\cite{khot2020text,karpas2022mrkl} by coupling the supervision for question decomposition to the supervision for performing the intermediate steps. Moreover, its non-modular nature does not allow using alternate symbolic reasoning engines in cases where they perform better than LMs.
Additionally, the model gets exposed to only a single set of in-context examples, selected based on their proximity to the complex question, which may not contain optimal supervision for the intermediate steps that need to be taken. 

We propose ``Successive Prompting'', where we iteratively decompose the complex question into the next simple question to answer, answer it, and then repeat until the complex question is answered (Figure~\ref{fig:sim_q}).  Each of these steps is performed with separate a query to the LM.  Since the decomposition and answering steps are performed separately, we can decouple the supervision of each step, providing two primary benefits.  First, when performing in-context learning, we get multiple opportunities to select different in-context examples, which can be tailored to the particular decomposition or answering step being performed, instead of selecting a single set of examples based only on the complex question.  Second, when fine-tuning (with or without in-context examples~\cite{chen-etal-2022-meta}), we can provide training examples for each step independently, so the model only has to learn to perform one step at a time. 

This decoupling additionally allows us to judiciously inject synthetic data into the learning process, e.g., to help the model answer a particular kind of simple question that it could not previously answer, or a new reasoning composition it did not know how to decompose. Because the steps are separate, we can isolate model failures and develop synthetic approaches to fill in the gaps.  It also allows us to replace the LM with other, purpose-built components to perform symbolic reasoning when appropriate~\cite{khot2020text,segal2019simple,jin2021mtmsn}.

We demonstrate the utility of successive prompting using a few-shot variant of the DROP dataset~\cite{dua2019drop}, selecting 300 examples for training (either fine-tuning or in-context example selection). These 300 examples are manually annotated with simple QA pairs as decompositions.  We find that performance of all models is quite low in this few-shot setting, so we develop a synthetic data generator that produces complex questions with their decompositions from semi-structured Wikipedia tables~\cite{yoran2021turning}.  This synthetic data provides not just complex question supervision, but also supervision for the intermediate steps. We augment this data with the 300 (complex) training examples and their decompositions from DROP. In this few-shot setting, our best performing successive prompting model shows a $\sim$5\% improvement in F1 when compared to state-of-the-art model on DROP. The code and data are available at \url{https://github.com/dDua/succesive_prompting}

\section{Decomposing Complex Questions}
\begin{figure*}
\begin{center}
\includegraphics[scale=0.68]{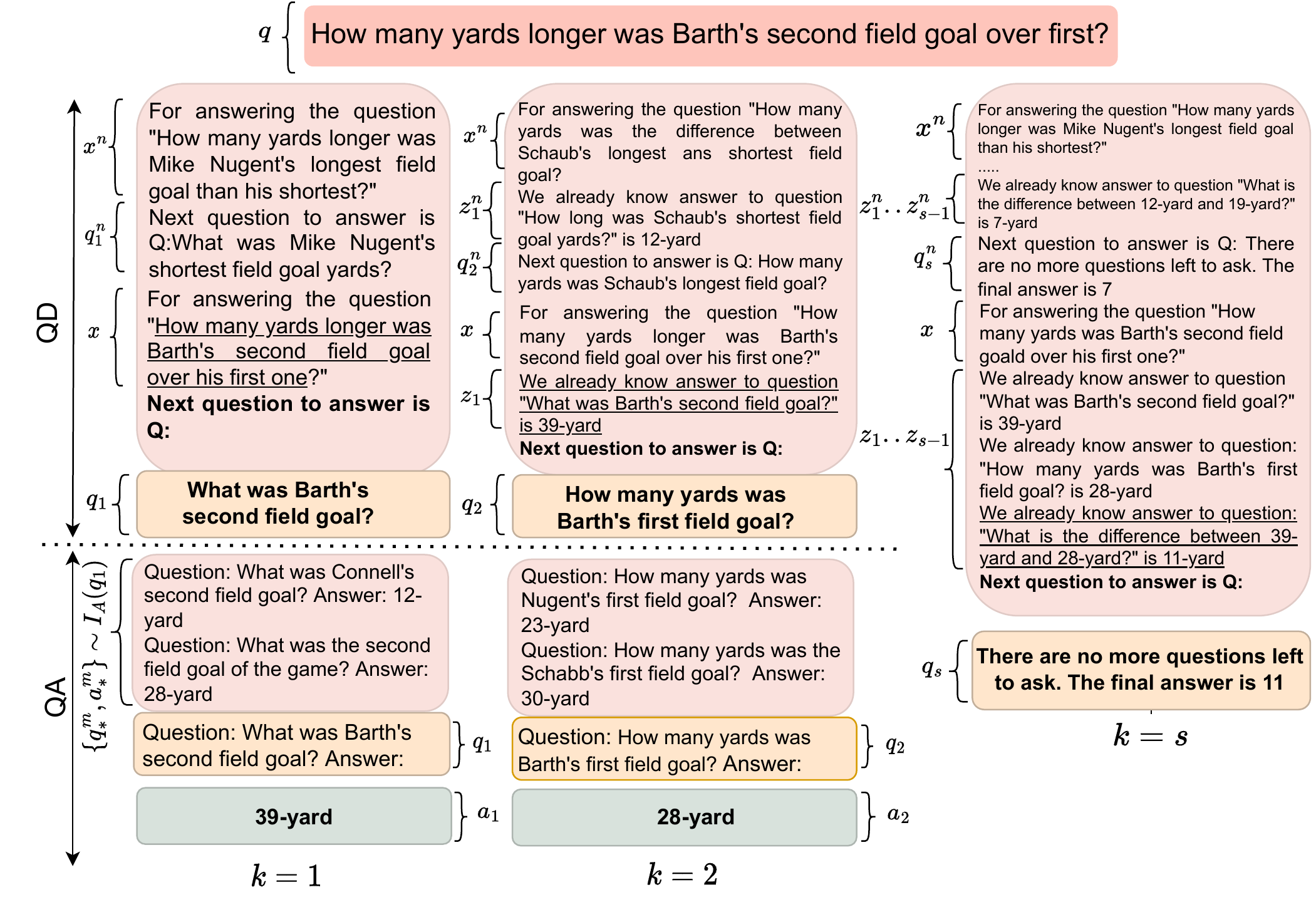}
\caption{A demonstration of successive prompting with in-context learning. The \hllightsalmon{selected examples for supervision} and \hldarksalmon{complex question to be answered} pre-pended with the context paragraph (omitted to simplify illustration) are encoded by the model to generate {\hllightbeige{question}} and  {\hllightgreen{answer}} at QD and QA stage respectively. During fine-tuning, only \hllightsalmon{training supervision} is used in an i.i.d manner for learning QD and QA models. }
\label{fig:step_by_step}
\end{center}
\end{figure*}

The goal of compositional question answering is to answer a complex question $q$ in the context of a passage $p$ (together denoted as ${x}$) by reasoning through latent sequential decisions $\mathbf{z} = z_1, z_2,...,z_s$ to reach the final answer, ${y}$. Many models have been proposed to accomplish this with varying amounts of supervision and interpretability.  
In prompting methods like Chain-of-Thought~\cite[CoT,][]{wei2022chain} the latent steps are supervised, interpretable sentences; in other models these latent steps might be a program~\cite{gupta2019neural,chen2019neural} or even just the (unsupervised) hidden states in the model~\cite{segal2019simple,andor2019giving}

We focus on models that take in-context examples and produce a discrete, language-encoded $\mathbf{z}$, with CoT being the primary exemplar.
We write the general form for CoT, given an input ${x}$, a language model encoder $\mathbb{L}$ and $N$ in-context examples obtained from querying an index $\mathcal{I}$---each containing a triplet of passage with complex question (${x^n}$), latent steps ($\mathbf{z^n}$) and final answer (${y^n}$)---as follows:
\begin{align*}
    {y}, {z} &\leftarrow \mathbb{L} ~\Big(x, \Set{\big(x^n, y^n, \mathbf{z}^n \big) \mid n \in [1,N] } \Big).
\end{align*}

\subsection{Successive prompting}
In successive prompting, we represent each latent step as a pair of simple question and answer, $z_k=(q_k,a_k)$ (see Figure~\ref{fig:sim_q} for example QA pairs) unlike CoT which represents each latent step as a declarative sentence. Moreover, CoT queries the index $\mathcal{I}$ for in-context examples and prompts the language model $\mathbb{L}$ for generating output only once. However, in successive prompting, we separate $\mathbf{z}$ into multiple question and answering steps, which gives us many opportunities to prompt $\mathbb{L}$, with potentially different in-context examples that are more tailored to the simple question at each step. It also enables us to re-encode the context given the intermediate state $z_k$, which can be useful in certain questions that need long chain referencing (e.g., the sort-count example in Figure~\ref{fig:synthetic_count}). We can write a general form for successive prompting as follows: 
\begin{align*}
    q_1 &\leftarrow \mathbb{L} ~\Big(x, \Set{ \big(x^n, q_1^n \big) \mid n \in [1,N] } \Big) \\
    a_1 &\leftarrow \mathbb{L} ~\Big(p, q_1, \Set{ \big(p_*^m, q_*^m, a_*^m \big) \mid m \in [1,M] } \Big) \\
\end{align*}
\begin{align*}
    q_2 &\leftarrow \mathbb{L} ~\Big(x, q_1, a_1, \Set{ \big(x^n, q_1^n, a_1^n, q_2^n \big) \mid n \in [1,N] } \Big) \\
    a_2 &\leftarrow \mathbb{L} ~\Big(p, q_2, \Set{ \big(p_*^m, q_*^m, a_*^m \big) \mid m \in [1,M] } \Big) \\
    \cdots \\
    y &\leftarrow \mathbb{L} ~\Big(x, \mathbf{z}, \Set{\big(x^n, y^n, \mathbf{z}^n \big) \mid n \in [1,N] } \Big)
\end{align*}
There are three kinds of model outputs in this general form: intermediate questions $q_k$, intermediate answers $a_k$, and the final answer $y$.  We refer to the first kind of output as \emph{question decomposition} (QD) and the second kind as \emph{question answering} (QA).  We treat final answer prediction as a special case of question decomposition, where the model decides that no more decomposition is necessary and outputs a final answer, so we iteratively alternate between question decomposition and question answering until the model terminates.

\subsection{Training paradigm}
\label{sec:training}
We have so far described successive prompting in a setting where only in-context examples are given, so no model training is performed.  However, successive prompting can also be used in conjuction with model fine-tuning, where each intermediate output is treated as a training example for $\mathbb{L}$. In this section, we first describe how in-context examples are selected at every step, followed by detailing how these examples are used for model fine-tuning.

\paragraph{In-context Learning}
During in-context learning, a small number of training examples are provided directly in the prompt that is given to a large LM, before the test input.  These examples are selected from an index based on their similarity with the test input. For successive prompting, we create two indices: $\mathcal{I}_D$, for looking-up relevant demonstrations for QD, and $\mathcal{I}_A$, for looking-up relevant demonstrations for QA. The index $\mathcal{I}_D$ contains partially decomposed chains at each step $k$, demonstrating the next question $q_k$ to be produced for every complex question in the training data. The index $\mathcal{I}_A$ contains all the simple QA pairs in the training data from all the complex questions.

In the QD stage, the index $\mathcal{I}_D$ is queried with the complex test question, $q$ and current step number, $k$, to select demonstrations regarding how to generate the next question for the held-out example. In the QA stage, the index $\mathcal{I}_A$ is queried with the simple question $q_k$ generated during QD to select relevant simple QA pairs. Figure~\ref{fig:step_by_step} shows a demonstration of how in-context learning is executed step-by-step in each stage until QD outputs the special phrase ``There are no more questions left to ask'', along with a final answer.

Successive prompting allows the QA stage to access simple questions derived from complex questions that would not have been retrieved by Chain-of-Thought prompting because on the surface they are not similar to the held-out complex question, even though they share similar sub-questions. 

\paragraph{Model Fine-tuning}
For model fine-tuning, we use T5~\cite{raffel2020exploring} based sequence-to-sequence models. Such models are typically trained with control codes in a multi-task setting~\cite{ma2021contrastive,rajagopal2022counterfactual} to switch between QD and QA tasks with shared model parameters. We adapt and extend the control codes introduced by text modular networks~\cite[TMNs,][]{khot2020text} for training with our synthetic data. TMNs are limited in terms of the operations they can handle as they do not go beyond first order reasoning. We use synthetically generated data, which allows us to deal with higher-order reasoning questions in DROP. 
Because we are fine-tuning the model, we can use special tokens to denote question decomposition and other separators, instead of the natural language prompts shown in Figure~\ref{fig:step_by_step}, though the content is the same.  The specific tokens used for each step are listed in Appendix A.

\paragraph{Specialized Modules} Successive prompting also allows us to use specialized sub-modules for solving different QA tasks because we no longer perform QD and QA in an end-to-end manner. Solving arithmetic operations like counting, difference, sorting, etc., can be challenging for language models. As a result, we follow \citet{khot2020text} and construct a simple mathematical sub-module for QA which parses the generated simple question for symbolic operation type and its arguments and then executes them in a deterministic way. If the generated simple question cannot be parsed as a mathematical operation, we apply the language model to solve it.


\section{Synthetic Dataset}
Any method that prompts LMs to produce intermediate reasoning steps to answer complex questions needs some amount of supervision for those reasoning steps. This kind of annotation can be expensive to collect and often requires expert knowledge.  Prior work has typically relied on a small handful of manually-written example decompositions.  We find that such small collections lead to very poor performance on a dataset as varied as DROP, even for large models.  

To mitigate these data issues, we propose a way to synthetically generate complex questions and their decompositions using semi-structured data which is easy to parse. We show that we can bootstrap model learning with this out-of-domain, synthetically generated data so it can adapt better when fine-tuned with limited in-domain supervision. 


\paragraph{Generation Process:}
Inspired by \citet{yoran2021turning}, we use semi-structured data from tables in English Wikipedia which are available in plenty.
We employ curated templates to convert the rows in the tables into paragraphs. We use single column headers to create first order simple questions and a combination of columns for higher order complex questions.  
We synthesize data for 10 simple operations: \texttt{COUNT}, \texttt{TOP(k)}, \texttt{BOTTOM(k)}, \texttt{FILTER}, \texttt{SUM}, \texttt{COMPARISON}, \texttt{DIFFERENCE}, \texttt{NEGATION}, \texttt{GATHER}, and \texttt{INTERSECTION}.

We generate higher order combinations of first-order operations, wherever possible. Figure~\ref{fig:synthetic_count} shows examples of higher order combinations of the atomic operation \texttt{COUNT} with a few other simple operations using Table~\ref{tab:wiki} as context. The complete list of all decompositions is provided in Appendix A. Depending on the model, we use either symbolic or natural language version of the arithmetic operations.  If we are using an LM to perform arithmetic operations, we output natural language; if we are using a separate symbolic reasoning engine, we output symbolic operations.  We generate approximately 141K total complex questions which result in 525K examples for QD and 257K examples for QA. See Appendix A for more dataset statistics. 


\begin{table}[tb]
\begin{center}
\small
\setlength{\tabcolsep}{2pt}
\begin{tabular}{lllcc}
 \toprule
Round & Date & Opponent & Venue & Attendance \\
\midrule
R2 1st Leg & 26 Sep 1990 & Walsall & A & 5,666 \\
QFR & 23 Oct 1990 & Liverpool & H & 18,246 \\
SF 1st Leg & 24 Feb 1991 & Sheffield Wed. & H & 14,074 \\
SF 2nd Leg & 27 Feb 1991 & Oxford United & A & 34,669 \\
QFR & 23 Jan 1991 & Portsmouth & A & 33,861 \\
\bottomrule
\end{tabular}
\end{center}
\caption{Example table from Wikipedia where rows become sentences and columns are used for question generation (used as context for Figure~\ref{fig:synthetic_count}).}
\label{tab:wiki}
\end{table}
\begin{figure*}


\small
\begin{tabular}{lp{12.5cm}}

\toprule
\bf Reasoning & {\bf Complex Question} and {\bf Decomposition (\textcolor{black}{Question} [\textcolor{violet}{Natural Language} or \textcolor{brown}{Symbolic}], \textcolor{teal}{Answer})} \\

  \midrule
 Count &   
 \begin{minipage}[t]{\linewidth}
 How many opponents were there?
 \begin{itemize}[noitemsep,topsep=0pt,leftmargin=5mm]
    \item  \textcolor{violet}{Q: What are all the opponents?}
    \textcolor{teal}{Ans: Walsall; Liverpool; Sheffield Wed.; Oxford United; Portsmouth} \\
    \item \textcolor{brown}{Q: count(Walsall; Portsmouth; Sheffield Wed.; Oxford United; Portsmouth)}
    ~ \textcolor{teal}{Ans: 5}
    \begin{itemize}[leftmargin=2mm]
         \item \textcolor{violet}{Q: How many items are in the list: Walsall, Liverpool, Sheffield Wed. and \\ Oxford United, Portsmouth?} \\
    \end{itemize}
    
  \end{itemize}
\end{minipage} \\
\multicolumn{2}{l}{\bf Higher order decompositions} \\
\addlinespace
Sort-Count &    
 \begin{minipage}[t]{\linewidth}
 Which venue had the most number of opponents?
\begin{itemize}[noitemsep,topsep=0pt,leftmargin=5mm]
    \item \textcolor{violet}{Q: What are all the venues?} ~ \textcolor{teal}{Ans: A; H} 
    \item \textcolor{violet}{Q: What are opponents when venue was A?}   ~ \textcolor{teal}{Ans: Walsall; Oxford United; Portsmouth} 
    \item \textcolor{brown}{Q: count(Walsall; Oxford United; Portsmouth)}  ~ \textcolor{teal}{Ans: 3} 
    \item \textcolor{violet}{Q: What are opponents when venue was H?} ~ \textcolor{teal}{Ans: Liverpool; Sheffield Wed.}
    \item \textcolor{brown}{Q: count(Liverpool; Sheffield Wed.)}  ~ \textcolor{teal}{Ans: 2} 

    \item \textcolor{brown}{Q: top(1, 2;3)} \textcolor{teal}{Ans: 3}
    \begin{itemize}
        \item \textcolor{violet}{Q: What is the largest value in: 2 and 3?}
    \end{itemize}
    \item \textcolor{violet}{Q: Which venue has 3 opponents?} \textcolor{teal} {Ans: A}
\end{itemize} 
\end{minipage} \\
\addlinespace
Comparison-Count &  
 \begin{minipage}[t]{\linewidth}
 Which round had more venues: SF 1st Leg or QFR??  
\begin{itemize}[noitemsep,topsep=0pt,leftmargin=5mm]
    \item \textcolor{violet}{Q: What are the rounds when venue was A?}  ~ \textcolor{teal}{Ans: R2 1st Left; SF 2nd Leg; QFR}
    \item \textcolor{brown}{count(R2 1st Left; SF 2nd Leg; QFR)}  ~ \textcolor{teal}{Ans: 3}
    \item \textcolor{violet}{Q: What are the rounds when venue was H?}  ~ \textcolor{teal}{Ans: QFR; SF 1st Leg}
    \item \textcolor{brown}{count(QFR; SF 1st Leg)}  ~ \textcolor{teal}{Ans: 2} 
    
    \item \textcolor{brown}{if\_then(1 > 2; SF 1st Leg; QFR)}  ~ \textcolor{teal}{Ans: QFR}
    \begin{itemize}[leftmargin=2mm]
        \item \textcolor{violet}{Q: If 1 > 2 then answer is SF 1st Leg else it is QFR}
    \end{itemize}
\end{itemize}
\end{minipage} \\
\bottomrule
  
\end{tabular}

\caption{Examples of \texttt{COUNT} operation and some of its higher order combinations, with \textcolor{violet}{natural language} and \textcolor{brown}{symbolic} decompositions of the complex question. Underneath the first instance of a \textcolor{brown}{symbolic} operation we show its corresponding \textcolor{violet}{natural language} version. See Table~\ref{tab:wiki} for the original table used to generate context and questions.}
\label{fig:synthetic_count}
\end{figure*}

\section{Experiments and Results}
The DROP dataset contains a variety of reasoning compositions which are not uniformly distributed. In order to get a fair representation of DROP examples, we first embed the examples using a sentence embedding method trained on the QQP dataset~\cite{reimers2019sentence}.  We then use cosine similarity to get the top-50 nearest neighbor questions for each training example. The connection graph between each training question to its neighbors is then used to obtain 300 questions that cover the majority of the training data, via the vertex cover algorithm. We manually annotate these 300 examples with decomposed QA pairs in the same format as our synthetic data (Figure~\ref{fig:synthetic_count}). For synthetic examples, since we know the reasoning types, we uniformly sample example demonstration from each reasoning type.

\subsection{In-context Learning}
\paragraph{Setup}
We use faiss\footnote{\url{https://ai.facebook.com/tools/faiss/}} index with the QQP-based sentence embedding~\cite{reimers2019sentence} for indexing all the questions. We use GPT-J (6B)\footnote{\url{https://github.com/EleutherAI}} which is the largest freely available model we could use with prompts containing 6 in-context examples. 

\paragraph{Results}
In Table~\ref{tab:inctx}, we compare performance of language models without any prompting (Standard), with chain-of-thought prompting (CoT) and successive prompting. We observe that successive prompting performs better than CoT by 3.5\% when only synthetic data is available, and 4.3\% better with synthetic data and 300 annotations from DROP. The best successive prompting version on the dev set (Synthetic+DROP) has a test set performance of 30.6\% F1. 
We also perform an ablation where the symbolic calculator is replaced by language model and observe that the performance drops by ~1.5\% F1. This further shows that modular approach is better over a single model that tries to solve all the tasks.

\begin{table}[tb]
\centering
    \small
    \begin{tabular}{lccc}
    \toprule
      & {\bf Syn-Only} & {\bf DROP-Only} & {\bf Syn+}\\
       &  & & {\bf DROP}\\
      \midrule
      Standard & 22.7 & 23.8 & 24.9\\
      CoT & 25.3 &  26.2 & 27.6\\
      Succ.(w/o calc.) & 27.2 & 29.3 & 29.9 \\
      Succ.(w/ calc.) & 28.8 & 30.8 & 31.9 \\
    \bottomrule
    \end{tabular}
    \caption{F1 Performance of in-context prompting on the DROP dev set with and without in-domain annotations.}
    \label{tab:inctx}
\end{table}

\subsection{Model Fine-tuning}

\paragraph{Setup}
We employ a shared question decomposition (QD) and answering model (QA) based on T5-large version of UnifiedQA~\cite{khashabi2020unifiedqa}, trained in a multi-task manner.  We use the format described in Appendix A for prompting UnifiedQA. For symbolic questions, we use a simple calculator that parses the operator and arguments in the generated question and executes the discrete operator on the detected arguments. 

To deter the model from learning incorrect steps, we use contrastive estimation~\cite{smith2005contrastive}. In particular, we first train the model for two epochs with cross-entropy loss while generating the output sequence (simple question or answer). Then we continue training by adding an auxiliary loss term which increases the likelihood of the intermediate sub-question that would produce a correct sub-answer at the cost of one that does not~\cite{dua2021learning}. We sample up to 3 negative samples at each step.  We use HuggingFace transformers\footnote{\url{https://github.com/huggingface/transformers}} to train our models, with a learning rate of 5e-5 and maximum input length of 768. 

Due to variance in the types of context tables present in Wikipedia, the synthetic dataset distribution is not uniform across different reasoning types. To have a balanced representation of questions pertaining to different reasoning types, we employ dynamic sampling~\cite{gottumukkala2020dynamic}, where at the beginning of each epoch we select 80,000 instances from across all reasoning types in proportion to the drop in their current performance with respect to previous epoch on held-out synthetic data. For the first epoch we sample in proportion to original the size of each reasoning type.
During inference, we use beam search with size 5 to generate decompositions, switching between QD and QA stages until QD reaches end of decomposition (``EOQ'') or maximum number of steps which we set as 10.

\paragraph{Baseline models} We compare against a number of different baselines, both symbolic and non-symbolic.  As non-symbolic baselines, we use UnifiedQA~\cite{khashabi2020unifiedqa}, which is pre-trained on a number of existing question answering datasets, and PReasM~\cite{yoran2021turning}, which is pre-trained on synthetically generated compositional QA pairs.  We also include a baseline with symbolic components, TASE~\cite{segal2019simple}.  This model (and others like it~\cite{jin2021mtmsn,andor2019giving}) are capable of performing a combination of continuous and discrete operations, which is essential for DROP. TASE does not require expressing decomposition in a specific grammar and can work with natural language. We chose this model as it is close to state of the art on the full DROP dataset and has publicly available code. 

\paragraph{Results}
In Table~\ref{tab:model_comp}, we use the DROP dev set to compare the performance of different symbolic and non-symbolic models in three settings: (1) using no training data from DROP (0-shot), (2) using only question-answer supervision from the 300 DROP examples, and (3) using both question-answer supervision and the decompositions for the 300 DROP examples.  In each of these settings, we can train the model with or without the synthetic data that we generated.  

We observe that our out-of-domain synthetic data universally improves model performance, and the improvement is most pronounced in TASE, nearing a 20\% absolute improvement.  Without synthetic data, PReasM is the best performing baseline, but TASE overtakes PReasM when synthetic data is available.  Additionally, and unsurprisingly, increasing the amount of supervision from 0-shot to complex QA pairs to decompositions universally improves model performance.

Finally, our method, which is a fine-tuned successive prompting model combined with a symbolic reasoning engine, achieves the best performance, giving an improvement of 5.4 F1 over the state-of-the-art model with similar supervision, i.e. TASE+Synthetic w/ decomp.
We follow the standard practice of using test set for only our final best performing model (SP w/ decomp). We observe that our best model with a test set performance of 50.2 F1 is better than the state-of-the-art model with similar supervision (45.1 F1) by 5.1\% F1.

\begin{table}[tb]
\centering
    \small
    \begin{tabular}{lccc}
    \toprule
      &   0-shot & {w/o decomp} & {w/ decomp} \\ 
     \midrule
     \multicolumn{4}{l}{\bf Non-symbolic} \\
      \emph{UnifiedQA} & 24.5 & 26.7 & 27.2 \\ 
      + Synthetic & 26.6 & 30.3 & 32.6  \\ 
      \emph{PReasM} & 24.9 & 34.6 & 37.5  \\ 
      + Synthetic & 30.2 & 36.2 & 38.1 \\ 
      \addlinespace
      \multicolumn{4}{l}{\bf Symbolic} \\
      \emph{TASE} & - & 26.1 & 27.6  \\ 
      + Synthetic & 27.3 & 44.1 & 45.9 \\ 
      \emph{Succ. Prompting} & 49.8 & - & 51.3 \\ 
    \bottomrule
    \end{tabular}
    \caption{F1 Performance of various model architectures on DROP dev-set pre-trained on synthetic data and further fine-tuned with 300 DROP examples.}
    \label{tab:model_comp}
\end{table}

Overall, methods that learn to decompose complex questions into simple QA pairs adapt well to complex questions in new domain with little (SP w/ decomp: 51.3 F1) to no in-domain supervision for decomposition (SP 0-shot: 49.8).
If we have limited complex QA supervision (without any decompositions), un-interpretable symbolic models result in the best performance (TASE + Synthetic w/o decomp: 44.1). This is because of two reasons. First, such models can capture domain specific answer priors which may result it decent held-out performance~\cite{dua2020benefits,agrawal2018don}. Second, depending on the context, sometimes it may not be straight-forward to decompose the complex questions into QA pairs.  



\subsection{In-context vs Fine-Tuning}
To understand the gap in performance between successive prompting with in-context learning and fine-tuning, we perform ablations across in-context and fine-tuned version of QD and QA modules. We observe that in-context learning is unable to do well on answering simple questions that result in a list of answers---which is especially important for DROP as symbolic aggregations are generally applied on a list of answers. On using a fine-tuned QA model we see an improvement of $\sim$10\% in F1 with an in-context QD model. Moreover, since the final answer performance is dependent on how well the QA model performs, using a better QD model (fine-tuned) does not help the overall performance much unless the QA model can handle the decompositions produced by the QD model.

\begin{table}[tb]
\centering
    \small
    \begin{tabular}{lcc}
    \toprule
      &  {QA: In-Context} & {QA: Fine-tuning} \\
     \midrule
      QD: In-Context & 30.8  & 40.3 \\
      QD: Fine-tuning & 31.4 & 51.3 \\
    \bottomrule
    \end{tabular}
    \caption{F1 with QD and QA modules from incontext learning and fine-tuning with only DROP annotations}
    \label{tab:abl}
\end{table}

\subsection{Qualitative Examples}
\begin{figure*}
\footnotesize
\begin{subfigure}[b]{\textwidth}
\begin{center}
\begin{tabular}{|M{2.8cm}|M{3.6cm}|M{3.8cm}|M{3.6cm}|}
     \hline
      {\bf Example} & {\bf Chain-of-Thought} &  {\bf Successive (In-context)} & {\bf Successive (Fine-Tuning)} \\
      \hline
      \textbf{Who threw the longest touchdown pass?}
      & The touchdown passes are 22-yard and eight-yard. Out of 22 and eight-yard, the longest touchdown pass is 22. Peyton manning threw the 22-yard touchdown pass. The answer is Peyton Manning. \emph{Answer} (\textcolor{olive}{\cmark}) & \begin{itemize}[noitemsep,leftmargin=2mm,topsep=1mm]
      \item What are all the touchdown passes? A: 22-yard, eight-yard
      \item  What is the largest value in: 22-yard, eight-yard? A: 22-yard
      \item Who threw the 22-yard touchdown pass? A: Peyton Manning.
      \item There are no more questions left to ask. The final answer is Peyton Manning. \emph{Answer}  (\textcolor{olive}{\cmark})
      \end{itemize} & \begin{itemize}[noitemsep,leftmargin=2mm,topsep=1mm]
      \item What are all the touchdown passes? A: 22-yard, eight-yard
      \item  max(22;eight) A: 22 
      \item Who threw the 22-yard touchdown pass? A: Peyton Manning. \emph{Answer} (\textcolor{olive}{\cmark})
      \end{itemize} \\
      \hline
      \textbf{In which two quarters did Stephen Gostkowski only kick one field goal?} \vspace{1mm}
      \scriptsize{Count-Argmax type decompositions were not covered in synthetic annotations}
      &  The first quarter is the only quarter when Stephen gostowski kick only one field goals. The count becomes 1. The second quarter is the only quarter when Stephen gostoswki only kick two field goals..... The answer is first quarter. \emph{Answer} (\textcolor{purple}{\xmark}) & \begin{itemize}[noitemsep,leftmargin=2mm,topsep=1mm]
      \item Which quarter did Gostowski score? A: first, second and fourth
      \item What are the field goals that Gostowski scored? A: 38, 24 and 21 
      \item There are no more question left to ask. The final answer is 38 and 24. \emph{Answer} (\textcolor{purple}{\xmark})
      \end{itemize} & \begin{itemize}[noitemsep,leftmargin=2mm,topsep=1mm]
      \item What are all the field goals when player was Stephen gostowski? A: 38-yard, 24 and 21-yard
      \item What were the quarters when 38-yard field goal was made? A: second \emph{Answer} (\textcolor{purple}{\xmark})
      \end{itemize} \\
      \hline
      \textbf{How many field goals did both teams kick in the first half?} 
      & The field goals in the game are 20, 24, 53, 30 and 40. The count becomes 7. \emph{Answer} (\textcolor{purple}{\xmark}) & \begin{itemize}[noitemsep,leftmargin=2mm,topsep=1mm]
      \item What are the field goals both teams kicked in the first half? A: 20, 24 and 53
      \item How many items are in the list 20, 24 and 53? A: 3 
      \item There are no more questions left to ask. The final answer is 3. \emph{Answer}  (\textcolor{olive}{\cmark})
      \end{itemize} & \begin{itemize}[noitemsep,leftmargin=2mm,topsep=1mm]
      \item What are all the field goals in the first half? A: 20; 24; 53
      \item count(20; 24; 53) A: 3 \emph{Answer} (\textcolor{olive}{\cmark})
      \end{itemize} \\
      \hline
      \textbf{How many of the personnel were not civil servants?}  & The personnel are 1,513. Since 1,513 is less than 3,183. The answer is not civil servants. \emph{Answer}(\textcolor{purple}{\xmark}) & \begin{itemize}[noitemsep,leftmargin=2mm,topsep=1mm]
        \item Q: How many of the personnel were not civil servants? A: 1,176 \item Q: What is the difference between 100 and 1176? A: 1111
        \item There are no more questions left to ask. The final answer is 1111. \emph{Answer} (\textcolor{purple}{\xmark})
      \end{itemize} & \begin{itemize}[noitemsep,leftmargin=2mm,topsep=1mm,topsep=1mm] 
        \item Q: How many personnel were there? A: 2,234
        \item Q: How many of the personnel were civil servants? A: 1,176
        \item Q: diff(2234; 1176) A: 1058
          \emph{Answer} (\textcolor{olive}{\cmark})
      \end{itemize} \\
     \hline
     \textbf{Q: Which port did the Korean immigrants leave first Chemulpo or Veracruz?}
      \footnotesize{Passage: In 1905, Korean immigrants departed from the port of Chemulpo.. After which they took a train to Veracruz. In the Veracruz port, another boat was taken to the port of Progreso.} 
        &  The Korean immigrants left Chemulpo. The answer is Chemulpo. Reasoning (\textcolor{purple}{\xmark}) Answer (\textcolor{olive}{\cmark}) & 
        
    \begin{itemize}[noitemsep,leftmargin=2mm,topsep=1mm] 
        \item Q: In what year did the Korean immigrants leave Chemulpo? A: 1905
        \item Q: Which port did the Korean immigrants leave first Chemulpo or Veracruz? A: Chemulpo
        \item Q: If Chemulpo is greater than Veracruz, answer is Chemulpo else it is Veracruz  A: Chemulpo.
          Reasoning (\textcolor{purple}{\xmark}) \emph{Answer} (\textcolor{olive}{\cmark})
      \end{itemize} 
        &
      \begin{itemize}[noitemsep,leftmargin=2mm,topsep=1mm] 
        \item Q: When did Korean immigrants leave Chemulpo? A: 1905
        \item Q: When did Korean immigrants leave Veracruz? A: 1905
        \item Q: if\_then(1905 $<$ 1905; Chemulpo; Veracruz)  A: Veracruz.
          \emph{Answer} (\textcolor{purple}{\xmark})
      \end{itemize}   \\
      \hline
\end{tabular}
\end{center}
\end{subfigure}
\caption{Generated decompositions depicting strength and weaknesses of Successive Prompting.}
\label{tab:train_ana}
\end{figure*}

To evaluate the correctness of decomposed QA pairs, we manually analyze a subset of predictions on the dev set with in-context (DROP-only) learning and model fine tuning (few shot). We do this by randomly sampling 50 correct predictions to determine how often the incorrect decompositions result in correct answer. We observe that QD stage has an accuracy of 88\% for in-context and 96\% for fine-tuned model. The incorrect decompositions are mainly because the decomposed question is identical to the original question. For instance, "Who made the longest field goal?" can sometimes be answered correctly without decomposing the question if the passage contains a single field goal mention. 


We also sample 50 incorrect predictions to ascertain the reason for incorrect predictions in both in-context and fine-tune setup.
We observe that the final predictions are incorrect due to three main categories of errors: incorrect QA model prediction, incorrect next question prediction (QD) and out-of-scope reasoning type. The QA model outputs incorrect answers to simple question 40\% and 22\% of the times for in-context and fine-tuned respectively. The second class of errors, due to incorrect decomposition, occur 30\% of the times for both in-context and fine-tuned. The final class of errors, due to compositional questions that are not covered by synthetically generated annotations, occur 28\% (in-context) and 46\% (fine-tune) of the times.


In Figure~\ref{tab:train_ana}, we show a few examples of correct and incorrect predictions and point out the strengths and weaknesses of successive prompting.
The main strength of successive prompting is that, by breaking down the question, we are able to get improved supervision for QA. 
As a result, it is able to correctly identify the goals kicked in the first half while answering the question ``How many field goals did both teams kick in the first half?", unlike CoT that returns goals for the entire game.

One of the limitations of in-context learning, when compared with fine-tuning (irrespective of the type of prompting), is that examples are chosen based on the question alone, overlooking the context. For instance, DROP has questions like ``How many people were not Germans, in terms of percentage?'' where we first need to answer ``How many people were Germans, in terms of percentage?" and then perform a negation operation (i.e, subtract from 100). The word ``not" influences the example lookup to choose decomposition that involves a negation even when the question being answered requires a different operation. 

A limitation of successive prompting is that it is sometimes challenging to decompose a question, especially when it involves implicit reasoning from the passage. For instance, for ``Which port did the Korean immigrants leave first Chemulpo or Veracruz?'', it is difficult to explicitly define a comparison style decomposition from the sentence, ``After which they took a train to Veracruz''.

\section{Related Work}
\paragraph{Prompting methods}
Prompting was introduced as a way to test the reasoning capabilities of large language models~\cite{brown2020language}. 
In follow-up works~\cite{schick2022few,chowdhery2022palm,marasovic2021few} prompting techniques have been used as a mechanism to supervise the model decision with few demonstrations as a conditioning context to guide its predictions on an unseen example. Works like Chain-of-Thought reasoning~\cite{wei2022chain,zelikman2022star} especially focus on compositional questions where they provide a chain of reasoning as demonstrations. 
In concurrent work, Least-to-Most prompting~\cite{zhou2022least} takes a similar view as ours to break down the problem into sub-problems. However, in Successive Prompting the question decomposition and answering stages are interleaved, unlike Least-to-Most where the problem is first reduced into sub-problem and then executed in a sequence. In our method, the next question prediction has access to previously answered sub-questions, which is useful in questions that need long chain referencing. 
Other contemporaneous works~\cite{press2022measuring,khot2022decomposed} use very large language models (more than twice the size we used) and show better few-shot generalization.  Works like~\citet{perez2021true} have shown the importance of having the right in-context examples for downstream performance leading to works that learn to retrieve relevant in-context examples~\cite{rubin2021learning}.

\paragraph{Non-symbolic methods}
Most non-symbolic methods are sequence-to-sequence models trained on a large amount of question answering data~\cite{khashabi2020unifiedqa,yoran2021turning}. 

\paragraph{Symbolic methods}
Neural module networks like approaches parse complex questions into a pre-specified grammar and learn neural components to handle symbolic mathematical operations~\cite{gupta2019neural,chen2019neural,nye2021show} which are recursively executed. State-of-the-art models on DROP, however, use a combination of BERT-based contextual models along with a calculator that performs discrete operations~\cite{andor2019giving,segal2019simple,hu2019multi}. Works like Text Modular networks~\cite{khot2020text} and MRKL~\cite{karpas2022mrkl} are closest to our work. However, they are limited in the terms of types of simple questions they can answer (single-span only) and the complexity of reasoning they can do (single-order only). TMNs, additionally, use a classifier that scores the generated chains module and filters out incorrect question decompositions, while we use contrastive estimation to learn a better question decomposer and as a result do not need a chain scorer.

\section{Conclusion}
We present a way to successively decompose complex questions into simple QA pairs, which allows for modular QD and QA systems that can be trained and queried independently. When performing in-context learning, we showed that successive prompting yields an improvement of 4.6 F1 over chain-of-thought prompting. When replacing just the in-context QA module with a fine-tuned one, which is adept at handling list type questions, we further improve the overall performance by 9.5 F1. We believe that modular systems that decompose and delegate tasks to the most appropriate model, whether that is a large LM or a tailored component, are more effective at solving complex tasks than trying to have a large LM solve the entire task on its own.  Successive prompting shows one way this decomposition and delegation can be done.

\section*{Acknowledgements}
We would like to thank Anthony Chen, Catarina Belem and the anonymous reviewers for the
discussions and feedback.
This material is based upon work sponsored in part by the DARPA MCS program under Contract No. N660011924033 with the United States Office Of Naval Research, in part by funding by AI2 and NSF IIS-1817183. We
would also like to thank Hasso Plattner Institute(HPI) for supporting the first author through UCI-HPI fellowship. The views in this work are of authors and not the sponsors.
\section*{Limitations}
We propose a way to decompose complex questions into interpretable simple QA pairs as latent steps that get successively asked and answered by large pretrained models. 
The notion of performing complex tasks by iteratively finding and then filling information needs is very general, but we have only shown the applicability of one specific version of this idea in one specific setting.  There are many potential challenges in applying successive prompting more broadly.  The biggest is that it requires at least some decomposition data, which may be hard or even impossible to obtain.  Some complex questions are not easily decomposed, and some domains can be very challenging to write synthetic data generators for.  We were able to generate synthetic data that covered most of the reasoning types in DROP, but other kinds of complex questions would not be covered by our generator (e.g., questions that require commonsense or causal reasoning). 

There is also significant difficulty in choosing a level of granularity for decomposition.  If a large pretrained model can directly answer a question as complex as ``What was Barth's second field goal?'', we should let the model answer the question instead of trying to decompose it further.  The right granularity for the decomposition thus depends on the capabilities of the underlying model, and those capabilities are rapidly changing as newer and larger pretrained models are released.  There is the possibility that newer model iterations will not need any decomposition to answer the complex questions covered by our synthetic data generator, making that generator obsolete.  However, it seems unlikely that pretrained models will be able to handle all complex scenarios in the near future, so the ideas of successive prompting and generating synthetic data to bridge reasoning gaps should still be applicable even when our particular application of them becomes obsolete.

This method also increases the computational requirements for answering complex questions, as instead of making one query to a large LM, successive prompting makes many queries to answer a single question.

\section*{Ethics Statement}
This work focuses on improving complex question answering with limited data. It uses existing training data and conventional methods of testing model performance. This work does not deal with any social impacts or biases in natural language processing systems.


\bibliographystyle{acl_natbib}
\bibliography{anthology,custom}
\appendix
\section{Appendix}
\subsection{Control codes for Model Fine-tuning}
To generate a simple question given complex question and previously generated latent steps, we append the input with control code ``QS:''
\begin{myfont}
\begin{center}
  \{$x^n$\} QI: $q_1^n$ A: $a_1^n$ $\cdots$ QI: $q_{k-1}^n$ A: $a_{k-1}^n$ QS:
  \end{center}
\end{myfont}

To answer the simple question we prompt the model again with only the simple question this time and control code ``A:'' to generate answer, $a_k^n$

\begin{myfont}
\begin{center}
  \{$p^n$\} QS: $q_k^n$ A:
  \end{center}
\end{myfont}

We alternate between the two stages till we reach the end of decomposition marker ``EOQ"

\begin{myfont}
\begin{center}
  \{$x^n$\} QI: $q_1^n$ A: $a_1^n$ $\cdots$ QI: $q_s^n$ A: $a_s^n$ QS: EOQ 
  \end{center}
\end{myfont}

\subsection{Synthetic Dataset Statistics}
\begin{table}[ht]
\centering
\begin{tabular}{lr}
 \toprule
Reasoning & Number of Examples \\
\midrule
Filter &  9634\\
Count &  11019 \\
Comparison & 12239 \\
Difference &  15433\\
Negation & 2020 \\
Intersection & 9089 \\
Sum & 18754 \\
Sort &  10663 \\
Sort-Filter & 5827  \\
Difference-Sort & 13872 \\
Sum-Sort & 3212 \\
Count-Filter & 7382 \\
Gather-Count & 1506 \\
Sum-Count &  4138 \\
Difference-Count & 4216 \\
Sort-Count & 6328 \\
Comparison-Count & 5998 \\
\hline
Total & 141,330 \\
\bottomrule
\end{tabular}
\caption{Synthetic dataset statistics}
\label{fig:synthetic_stats}
\end{table}


\begin{figure*}
\begin{subfigure}[b]{\textwidth}
\begin{center}
\begin{tabular}{p{2cm}p{3cm}p{3.5cm}p{1cm}p{2cm}}
 \toprule
Round & Date & Opponent & Venue & Attendance \\
\midrule
R2 1st Leg & 26 September 1990 & Walsall & A & 5,666 \\
R2 2nd Leg & 10 October 1990 & Portsmouth & H & 10,037 \\
QFR & 23 October 1990 & Liverpool & H & 18,246 \\
SF 1st Leg & 24 February 1991 & Sheffield Wednesday & H & 14,074 \\
SF 2nd Leg & 27 February 1991 & Oxford United & A & 34,669 \\
QFR & 23 January 1991 & Walsall & A & 33,861 \\

\bottomrule
\end{tabular}
\end{center}
\caption{Example wikitable where rows become sentences and columns are used for question generation}

\label{fig:wikitable}
\end{subfigure}
\vspace{0.1cm}
\begin{subfigure}[b]{\textwidth}
\begin{tabular}{p{1cm}p{4.2cm}p{7.5cm}}
\toprule
\multicolumn{1}{>{\centering\arraybackslash}m{2.5cm}} {Reasoning} & \multicolumn{1}{>{\centering\arraybackslash}m{4cm}} {\textcolor{magenta}{Complex Question}} & \multicolumn{1}{>{\centering\arraybackslash}m{7.5cm}} {Decomposition (\textcolor{brown}{Question}, \textcolor{teal}{Answer})} \\
 \hline
 Filter & What are the opponents when date was later than 21 January 1991 and attendance was less than 20000? & \begin{minipage}[t]{\linewidth}
 \begin{itemize}[noitemsep,topsep=0pt]
    \item \textcolor{brown}{Q: What are the opponents when date was later than 21 January 1991?} \textcolor{teal}{A: Sheffiedl Wednesday; Oxford United; Walsall }
    \item \textcolor{brown}{Q: Out of Sheffield Wednesday,  Oxford United and Walsall, which opponents have attendance less than 20000?} \textcolor{teal}{A: Sheffield Wednesday}
\end{itemize}
\end{minipage} \\
  \hline
 Count & How many opponents were there? & 
 \begin{minipage}[t]{\linewidth}
 \begin{itemize}[noitemsep,topsep=0pt]
    \item  \textcolor{brown}{Q: What are all the opponents?} \textcolor{teal}{A: Walsall; Portsmouth; Liverpool; Sheffield Wednesday; Oxford United}
    \item \textcolor{brown}{Q: count(Walsall; Portsmouth; Liverpool; Sheffield Wednesday; Oxford United)} \textcolor{teal}{A: 5}
\end{itemize}
\end{minipage} \\
 \hline
 Comparison & What round had a higher attendance: SF 2nd Leg or QFR? &  \begin{minipage}[t]{\linewidth}
    \begin{itemize}[nosep,after=\strut]
        \item \textcolor{brown}{Q: What was the attendance when round was SF 2nd Leg?} \textcolor{teal}{A: 34,669}
        \item \textcolor{brown}{Q: What was the attendance when round was QFR?} \textcolor{teal}{A: 33,861} 
        \item \textcolor{brown}{if\_then(34,669 > 33,861; SF 2nd Leg; QFR)} \textcolor{teal}{A: SF 2nd Leg}
    \end{itemize}
    \end{minipage} \\
 \hline
 Difference & What is the difference between attendances when the opponent was Oxford United and Portsmouth?  & \begin{minipage}[t]{\linewidth}
    \begin{itemize}[nosep,after=\strut]
        \item \textcolor{brown}{Q: What was the attendance when opponent was Oxford United?} \textcolor{teal}{A: 34,669} 
        \item \textcolor{brown}{Q: What was the attendance when opponent was Portsmouth?} \textcolor{teal}{A: 10,037}
        \item \textcolor{brown}{diff(34669; 10037)}  \textcolor{teal}{A: 24632}
        \begin{itemize}
        \item \textcolor{olive}{Q: What is the difference between 34669 and 24632?}
        \end{itemize}
    \end{itemize}
    \end{minipage}  \\
 \hline
 Sum & What were the total attendances when opponents were Walsall and Oxford United? & 
 \begin{minipage}[t]{\linewidth}
    \begin{itemize}[nosep,after=\strut]
        \item \textcolor{brown}{Q: What was the attendance when opponent was Walsall?} \textcolor{teal}{A: 5,666; 33,861} 
        \item \textcolor{brown}{Q: What was the attendance when opponent was Oxford United?}  \textcolor{teal}{A: 34,669}
        \item \textcolor{brown}{sum(5666; 33861; 34669)} \textcolor{teal}{A: 74196}
    \end{itemize}
    \end{minipage}  \\
  \hline
   Sort & Which opponent had the second highest attendance? & 
   \begin{minipage}[t]{\linewidth}
    \begin{itemize}[nosep,after=\strut]
        \item \textcolor{brown}{Q: What are all the attendances?} \textcolor{teal}{A: 5,666; 10,037; 14,074; 34,669; 33,861} 
        \item \textcolor{brown}{Q: top(2, 5,666;10,037;14,074;34,669;33,861} \textcolor{teal}{A: 33,861}
        \item \textcolor{brown}{What was the opponent when attendance was 33,861?} \textcolor{teal}{A: Walsall}
    \end{itemize}
    \end{minipage} \\

\bottomrule
 
\end{tabular}
\caption{Single order decompositions}
\end{subfigure}
\label{fig:synthetic}
\end{figure*}

\begin{figure*} \ContinuedFloat
\begin{subfigure}[b]{\textwidth}
\begin{tabular}{p{1.5cm}p{4.2cm}p{8cm}}
\toprule
\multicolumn{1}{>{\centering\arraybackslash}m{1cm}} {Reasoning} & \multicolumn{1}{>{\centering\arraybackslash}m{4.2cm}} {Complex Question} & \multicolumn{1}{>{\centering\arraybackslash}m{7.5cm}} {Decomposition (\textcolor{brown}{Question}, \textcolor{teal}{Answer})} \\
 \hline
 \multicolumn{3}{c}{\bf Higher-order combinations with SORT} \\
 \hline
 Sort-Filter &  Which opponent had the third lowest attendance after 1 January 1991? & \begin{minipage}[t]{\linewidth}
    \begin{itemize}[nosep,after=\strut]
        \item \textcolor{brown}{Q: What are all the attendances after 1 January 1991?} \textcolor{teal}{A: 14,074; 34,669; 33,861} 
        \item \textcolor{brown}{Q: bottom(3, 14,074;34,669;33,861} \textcolor{teal}{A: 14,074}
        \item \textcolor{brown}{What was the opponent when attendance was 14,074} \textcolor{teal}{A: Sheffield Wednesday}
    \end{itemize}
    \end{minipage} \\
  \hline

Difference-Sort & What is the difference between the second highest attendance and lowest attendance? & \begin{minipage}[t]{\linewidth}
    \begin{itemize}[nosep,after=\strut]
        \item \textcolor{brown}{Q: What are all the attendances?}  \textcolor{teal}{A: 5,666;14,074;18,246;14,074;34,669;33,861}
        \item \textcolor{brown}{Q: top(2,  5,666; 14,074; 18,246; 14,074; 34,669; 33,861}  \textcolor{teal}{A: 33861}
        \item \textcolor{brown}{Q: bottom(1, 5,666; 14,074; 18,246; 14,074; 34,669; 33,861)}  \textcolor{teal}{A:5666}
        \begin{itemize}
            \item \textcolor{olive}{Q: What is the smallest value in: 33861 and 5666?}
        \end{itemize}
        \item \textcolor{brown}{diff(33861;5666)} \textcolor{teal}{A: 28195}
    \end{itemize}
    \end{minipage} \\ \hline 
Sum-Sort & What was the total of highest attendance and third lowest attendance? & \begin{minipage}[t]{\linewidth}
    \begin{itemize}[nosep,after=\strut]
        \item \textcolor{brown}{Q: What are all the attendances?} \textcolor{teal}{A: 5,666; 14,074;18,246;14,074;34,669;33,861}
        \item \textcolor{brown}{Q: top(1,  5,666; 14,074; 18,246; 14,074; 34,669; 33,861}  \textcolor{teal}{A: 34669}
        \item \textcolor{brown}{Q: bottom(3, 5,666; 14,074; 18,246; 14,074; 34,669; 33,861}  \textcolor{teal}{A:18246}
        
        \item \textcolor{brown}{sum(33861;5666)}  \textcolor{teal}{A: 52915}
        \begin{itemize}
            \item \textcolor{olive}{Q: What is the sum of 33861 and 5666?}
        \end{itemize}
    \end{itemize}
    \end{minipage} \\
\hline 
 \multicolumn{3}{c}{\bf (Additional) Higher-order combinations with COUNT } \\
 \hline
 Count-Filter & How many rounds had venue as A and attendance greater than 30000? & 
 \begin{minipage}[t]{\linewidth}
\begin{itemize}[noitemsep,topsep=0pt]
    \item \textcolor{brown}{Q: What rounds had venue as A?}  \textcolor{teal}{A: R2 1st Left; SF 2nd Leg; QFR}
    \item \textcolor{brown}{Q: Out of rounds R2 1st Left, SF 2nd Leg and QFR, which had attendance greater than 30000? }  \textcolor{teal}{A: SF 2nd Leg; QFR}
    \item \textcolor{brown}{Q: count(SF 2nd Leg; QFR)}  \textcolor{teal}{A: 2}
    
\end{itemize}
\end{minipage} \\
\hline 
 Gather-Count & How many opponents were there for each of venue: A and H? & \begin{minipage}[t]{\linewidth}
    \begin{itemize}[nosep,after=\strut]
        \item \textcolor{brown}{Q: What are the opponents when venue was A?}  \textcolor{teal}{A: Walsall; Oxford United}
        \item \textcolor{brown}{Q: count(Walsall; Oxford United)}  \textcolor{teal}{A: 2}
        \item \textcolor{brown}{Q: What are the opponents when venue was H?}  \textcolor{teal}{A: Portsmouth; Liverpool; Sheffield Wednesday United}
        \item \textcolor{brown}{Q: count(Portsmouth; Liverpool; Sheffield Wednesday United)}  \textcolor{teal}{A: 3}
        \item \textcolor{brown}{Q: gather(2;3)}  \textcolor{teal}{A: 2 and 3}
    \end{itemize}
    \end{minipage}\\
 \hline

\end{tabular}
\caption{Higher order decompositions}
\end{subfigure}

\caption{Examples of decompositions with wikitables}
\label{fig:synthetic}
\end{figure*}

\begin{figure*} \ContinuedFloat
\begin{subfigure}[b]{\textwidth}
\begin{tabular}{p{1.5cm}p{4.2cm}p{8cm}}
\toprule
\multicolumn{1}{>{\centering\arraybackslash}m{1.5cm}} {Reasoning} & \multicolumn{1}{>{\centering\arraybackslash}m{4cm}} {Complex Question} & \multicolumn{1}{>{\centering\arraybackslash}m{8cm}} {Decomposition (\textcolor{brown}{Question}, \textcolor{teal}{Answer})} \\
 \hline
 \multicolumn{3}{c}{\bf Additional Higher-order combinations with COUNT } \\
 \hline
 Sum-Count & What are the total number of opponents when venue were A and H? & \begin{minipage}[t]{\linewidth}
    \begin{itemize}[nosep,after=\strut]
        \item \textcolor{brown}{Q: What are the opponents when venue was A?} \textcolor{teal}{A: Walsall; Oxford United}
        \item \textcolor{brown}{Q: count(Walsall; Oxford United)} \textcolor{teal}{A: 2}
        \item \textcolor{brown}{Q: What are the opponents when venue was H?} \textcolor{teal}{A: Portsmouth; Liverpool; Sheffield Wednesday United}
        \item \textcolor{brown}{Q: count(Portsmouth; Liverpool; Sheffield Wednesday United)} \textcolor{teal}{A: 3}
        \item \textcolor{brown}{Q: sum(2;3)} \textcolor{teal}{A: 5}
    \end{itemize}
    \end{minipage}\\
 \hline
    Difference-Count &   What is the difference between number of rounds when venue was A and H? &
 \begin{minipage}[t]{\linewidth}
\begin{itemize}[noitemsep,topsep=0pt,leftmargin=5mm]
    \item \textcolor{brown}{Q: What are the venues when round was SF 1st Leg?} \textcolor{teal}{Ans: H}
    \item \textcolor{brown}{Q: count(A)}  \textcolor{teal}{A: 1}
    \item \textcolor{brown}{What are the venues when round was QFR?} \textcolor{teal}{A: H;A}
    \item \textcolor{brown}{Q: count(H;A)} \textcolor{teal}{A: 2}
    \item \textcolor{brown}{diff(3; 2)}  \textcolor{teal}{Ans: 1}
    
\end{itemize}
\end{minipage}  \\
 
 \hline

\end{tabular}
\caption{Higher order decompositions}
\end{subfigure}

\caption{Examples of decompositions with wikitables}
\label{fig:synthetic}
\end{figure*}


\end{document}